\documentclass[conference]{IEEEtran}
\IEEEoverridecommandlockouts
\usepackage{cite}
\usepackage{amsmath,amssymb,amsfonts}
\usepackage{algorithm}
\usepackage{algorithmic}
\usepackage{graphicx}
\usepackage{textcomp}
\usepackage{xcolor}
\usepackage{booktabs}
\usepackage{multirow}
\usepackage{url}

\def\BibTeX{{\rm B\kern-.05em{\sc i\kern-.025em b}\kern-.08em
    T\kern-.1667em\lower.7ex\hbox{E}\kern-.125emX}}

\begin{document}

\title{Tail-Aware HiFloat4: W4A4 Post-Training Quantization for Wan2.2%
\thanks{Zhanfeng Feng and Shuai Guo contributed equally to this work.}}

\author{
\IEEEauthorblockN{
Zhanfeng Feng\textsuperscript{*},
Shuai Guo\textsuperscript{*},
Xin Di,
Long Peng,
Yang Cao,
Zhengjun Zha
}
\IEEEauthorblockA{
\textit{University of Science and Technology of China}\\
Hefei, China\\
\{xiaobigfeng, guoshuai676, dx9826, longp2001\}@mail.ustc.edu.cn, \{forrest, zhazj\}@ustc.edu.cn
}
}

\maketitle

\begin{abstract}
This report describes Tail-Aware HiFloat4, our submission to the low-bit text-to-video generation quantization challenge. Our method adapts the public ViDiT-Q post-training quantization pipeline to Wan2.2 under the HiFloat4 numerical format. We quantize the main linear layers in both Wan2.2 transformer modules with W4A4 HiFloat4 fake quantization, keep numerically sensitive boundary modules in high precision, and introduce an activation-tail-aware percentile calibration module for channel-mask construction. Together with compact PTQ-state restoration, this design reduces the influence of rare calibration outliers while keeping the runtime HiFloat4 arithmetic and sampling pipeline unchanged.
\end{abstract}

\begin{IEEEkeywords}
post-training quantization, HiFloat4, W4A4, diffusion transformer, text-to-video generation, Wan2.2, tail-aware calibration, percentile calibration
\end{IEEEkeywords}

\section{Introduction}
Diffusion models have become an important family of visual generators~\cite{ddpm,ldm}, and diffusion transformers further improve scalability for high-resolution image and video synthesis~\cite{dit,makeavideo,cogvideo,imagenvideo,hunyuanvideo,wan}. Their quality is largely supported by large transformer blocks, wide feed-forward networks, and attention projections that are repeatedly evaluated over denoising timesteps. This makes inference memory-intensive and computationally expensive. The cost is especially visible in video generation, where temporal length and spatial resolution multiply the number of latent tokens processed by the model.

Low-bit inference is therefore attractive for deployment. Post-training quantization (PTQ) is particularly relevant because it does not require retraining a large generative model. However, aggressive W4A4 quantization is difficult for diffusion transformers. Activations vary across denoising timesteps, spatial and temporal tokens, prompt conditions, and classifier-free guidance branches. A calibration statistic that is too conservative can preserve rare outliers but waste the limited 4-bit representation range on values that appear only occasionally. Conversely, overly aggressive clipping may damage semantic fidelity and temporal consistency.

The ICME 2026 low-bit large model quantization challenge focuses on this practical regime. In the HiFloat4 track, the model must use the provided 4-bit floating-point format and keep at most a small number of transformer blocks in high precision. Our solution follows the public ViDiT-Q PTQ workflow for diffusion transformers~\cite{viditq}, but adapts it to the Wan2.2 challenge pipeline. Wan2.2-I2V-A14B~\cite{wan22_hf} contains two transformer modules and is evaluated through an image-to-video interface. Since the task is prompt driven, we keep a fixed placeholder image condition for both calibration and inference, so that the comparison isolates the effect of quantization.

This paper presents Tail-Aware HiFloat4, a concise W4A4 HiFloat4 PTQ system for Wan2.2. The system has three design goals. First, it should obey the challenge numerical format by using HiFloat4 for both weights and activations in the target linear layers. Second, it should be stable enough for the dual-transformer Wan2.2 backbone by retaining boundary modules, such as embedding and output projection layers, in high precision. Third, it should make the calibration stage less sensitive to activation tails through percentile-based statistics.

The contributions are summarized as follows:
\begin{itemize}
\item We adapt a ViDiT-Q-style PTQ pipeline to Wan2.2-I2V-A14B and quantize the main linear layers in both Wan2.2 transformer modules with HiFloat4 W4A4 fake quantization.
\item We introduce tail-aware percentile activation statistics for channel-mask construction, replacing a hard maximum statistic in the SmoothQuant-style balancing path.
\item We provide a compact PTQ checkpoint format that stores inference-time quantization deltas rather than duplicating the full BF16 transformer weights.
\item We evaluate the resulting W4A4 system under matched Wan2.2 generation settings and report the main quality trade-offs across imaging quality, aesthetic quality, consistency, and motion smoothness metrics.
\end{itemize}

\section{Related Work}
\subsection{Post-Training Quantization}
Post-training quantization compresses pretrained neural networks without the cost of full retraining. Early deployment-oriented work showed that quantized integer inference can substantially reduce memory traffic and arithmetic cost for convolutional models~\cite{jacob2018quantization}. Subsequent PTQ methods improved calibration and rounding behavior, including adaptive rounding of pretrained weights~\cite{adaround} and layer-wise reconstruction for very large transformers~\cite{gptq}. Transformer PTQ has also studied activation outliers and mixed numerical paths. Dettmers et al.~\cite{llmint8} isolate outlier features for large transformers, while ZeroQuant combines group-wise quantization and distillation for large transformers~\cite{zeroquant}. SmoothQuant~\cite{smoothquant} transfers part of the activation quantization difficulty into the weights through per-channel scaling, AWQ~\cite{awq} further emphasizes activation-aware protection of salient weights, and QuaRot~\cite{quarot} removes outliers through equivalent rotations. Our method follows the activation-aware scaling viewpoint, but focuses on how the activation statistic should be estimated for W4A4 video generation under a fixed HiFloat4 format.

\subsection{Quantization for Diffusion Transformers}
Diffusion models differ from standard discriminative networks because the same network is evaluated over many denoising timesteps and conditioning states. PTQ4DM~\cite{ptq4dm} and Q-Diffusion~\cite{qdiffusion} show that timestep-dependent activation distributions make diffusion quantization substantially different from standard image classification PTQ. EfficientDM~\cite{efficientdm} studies efficient quantization-aware adaptation for low-bit diffusion models. More recent work focuses on transformer-based generators: PTQ4DiT~\cite{ptq4dit}, Q-DiT~\cite{qdit}, and VQ4DiT~\cite{vq4dit} study post-training quantization for diffusion transformers, where attention and feed-forward projections dominate the arithmetic. ViDiT-Q~\cite{viditq} further identifies multiple axes of activation variation, including timestep-wise, token-wise, condition-wise, and channel-wise variation. It combines calibration, channel balancing, optional rotation, and mixed precision for sensitive modules. We use ViDiT-Q as the public PTQ foundation, but target the Wan2.2 challenge setting and the HiFloat4 format.

\subsection{Video Generation Evaluation}
Text-to-video generation has evolved from early transformer and diffusion systems to larger open video foundation models~\cite{makeavideo,cogvideo,imagenvideo,hunyuanvideo,wan}. Its quality is multi-dimensional: a quantized model can maintain frame-level aesthetic quality while degrading subject identity, prompt alignment, or temporal coherence. VBench~\cite{vbench} evaluates video generation with dimensions such as imaging quality, aesthetic quality, overall consistency, subject consistency, and motion smoothness. These metrics are useful for diagnosing low-bit video generation because the failure modes of W4A4 quantization are not captured by a single scalar score.

\subsection{HiFloat4 Format}
HiFloat4 is a 4-bit floating-point-oriented numerical format for low-bit inference~\cite{hifloat4}. In the challenge setting, the numerical format is fixed, so the method should not redefine the representation. We therefore treat HiFloat4 as a prescribed quantize-dequantize map for both weights and activations. The proposed percentile calibration only changes the PTQ statistics used to prepare channel masks; it does not change the HiFloat4 arithmetic.

\section{Preliminaries}
\subsection{PTQ Notation}
We consider post-training quantization of a pretrained generator $f_\theta$ without updating the original parameters by gradient training. For a tensor $z$, a generic fake-quantization operator can be written as
\begin{equation}
    \mathcal{D}(\mathcal{Q}(z; \Delta, \mathcal{C}); \Delta),
\end{equation}
where $\mathcal{Q}$ maps values to a low-precision code set $\mathcal{C}$ under scale or format parameter $\Delta$, and $\mathcal{D}$ maps the low-precision code back to floating point for simulation. Integer PTQ typically chooses a uniform code set with explicit scale and zero point. In this work, the code set and dequantization behavior are determined by the HiFloat4 toolkit. We therefore write the weight and activation paths as
\begin{equation}
    \hat{W}=\operatorname{HiF4}(W), \quad
    \hat{x}=\operatorname{HiF4}(x),
\end{equation}
where $\operatorname{HiF4}(\cdot)$ denotes the released quantize-dequantize operator. The PTQ algorithm is responsible for preparing transformed tensors that are easier for this fixed operator to represent.

\subsection{Channel Balancing}
Channel balancing uses a diagonal mask to redistribute dynamic range between activations and weights before quantization. For a linear layer, the floating-point computation is unchanged by
\begin{equation}
    xW^T = (xM)(WM^{-1})^T,
    \quad M=\operatorname{diag}(m).
\end{equation}
After quantization, however, the choice of $m$ changes the approximation error because $\operatorname{HiF4}(xM)$ and $\operatorname{HiF4}(WM^{-1})$ have different effective ranges. SmoothQuant-style methods construct $m$ from weight and activation magnitudes. This makes calibration statistics central to W4A4 performance: an activation estimate that over-emphasizes rare outliers can make the common activation range too coarse, while an estimate that clips too aggressively can remove important semantic signal.

\subsection{Wan2.2 Challenge Setting}
Wan2.2-I2V-A14B~\cite{wan22_hf} is an image-to-video model with two transformer modules. In the challenge protocol, text prompts provide the semantic condition and a fixed placeholder image provides the required image condition. The target layers are linear projections inside the transformer modules. The method does not change the scheduler, prompt processing, video resolution, number of frames, or HiFloat4 arithmetic. This separation is important for camera-ready reproducibility: differences between BF16 and W4A4 runs should come from quantization and PTQ state restoration, not from a changed sampling protocol.

\section{Method}
\subsection{Overview}
Our pipeline consists of calibration, PTQ state construction, and quantized inference. Calibration runs the BF16 Wan2.2 model on a small prompt set and estimates input activation statistics for linear layers in both transformer modules. PTQ replaces the selected linear projections with HiFloat4 quantized counterparts, constructs per-channel masks from the collected statistics, and forms a compact quantization state. Quantized inference applies this state to the original Wan2.2 checkpoint and generates videos with the same sampling configuration.

\begin{figure*}[t]
\centering
\includegraphics[width=0.98\textwidth]{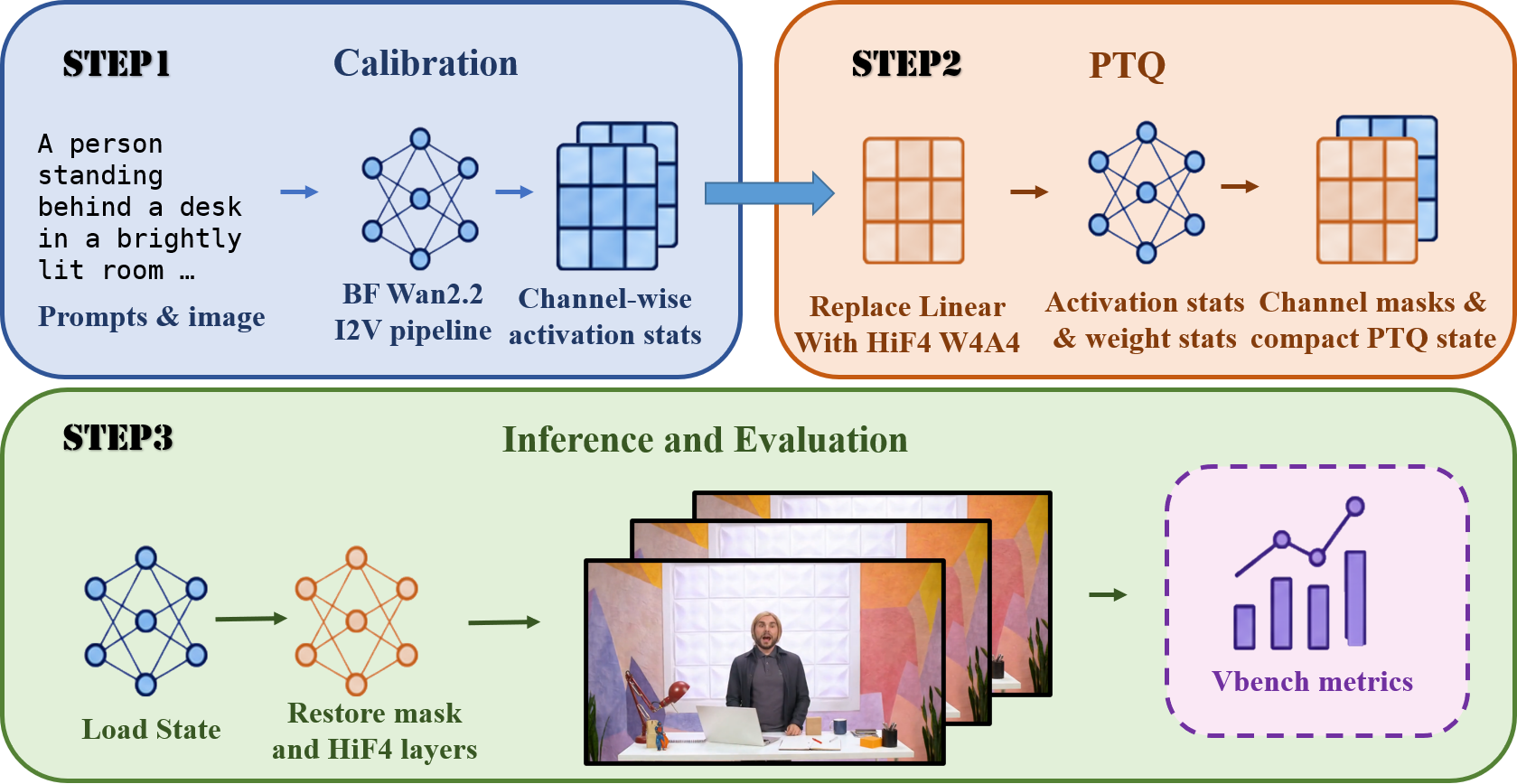}
\caption{Pipeline of the proposed Tail-Aware HiFloat4 W4A4 PTQ system for Wan2.2. Calibration collects activation statistics from the BF16 model, PTQ builds tail-aware percentile-calibrated channel masks and compact states, and inference restores those states while keeping the base sampler unchanged.}
\label{fig:pipeline}
\end{figure*}

Fig.~\ref{fig:pipeline} shows the procedure. The calibration and inference interfaces both use the same fixed placeholder image condition because the Wan2.2 generator follows an image-to-video formulation. Consequently, the BF16 and W4A4 runs share the same sampling protocol, and the remaining differences are attributable to quantization and prompt variation. Algorithm~\ref{alg:ptq} gives the layer-wise PTQ procedure used to build the compact state.

\begin{algorithm}[t]
\caption{Tail-Aware Percentile-Calibrated HiF4 PTQ}
\label{alg:ptq}
\footnotesize
\textbf{Input:} base model $f_\theta$, calibration prompts $\mathcal{P}$, percentile $p$, balancing coefficient $\alpha$\\
\textbf{Output:} compact PTQ state $\mathcal{S}$
\begin{algorithmic}[1]
\FOR{each target linear layer $\ell$}
    \STATE collect input activations $x_\ell$ on $\mathcal{P}$
    \STATE $a_i \leftarrow Q_p(\{|x_{\ell,j,i}|\}_j)$
    \STATE $w_i \leftarrow \max_o |W_{\ell,o,i}|$
    \STATE $m_i \leftarrow w_i^\alpha/(a_i+\epsilon)^{1-\alpha}$
    \STATE $\tilde{W}_\ell \leftarrow W_\ell \operatorname{diag}(m)^{-1}$
    \STATE $\hat{W}_\ell \leftarrow \operatorname{HiF4}(\tilde{W}_\ell)$
    \STATE store the layer mask and quantization state in $\mathcal{S}$
\ENDFOR
\end{algorithmic}
\end{algorithm}

\subsection{HiFloat4 Quantized Linear Layers}
Let a floating-point linear layer be
\begin{equation}
    y = x W^T + b,
\end{equation}
where $x$ is the input activation, $W$ is the weight matrix, and $b$ is the bias. For each selected projection, the weight path applies HiFloat4 quantize-dequantize to the channel-balanced weight, while the activation path applies the same numerical format to the online input activation. The quantization axes follow the natural matrix dimensions of the linear operation: weights are quantized over output channels and activations over feature channels.

Wan2.2-I2V-A14B contains two transformer modules, and we apply the same conversion rule to both. Matched attention projections and feed-forward projections are quantized. Normalization layers, embedding-related layers, patch embedding, final projection, and output-head style modules remain in BF16 for stability. The configuration also supports retaining at most two transformer blocks in high precision across both transformer modules, consistent with the HiFloat4 challenge rule. In the submitted configuration, no whole transformer block is retained in high precision.

\subsection{Tail-Aware Percentile Calibration}
For each target linear layer, hooks collect the absolute value of the input activations. Let $x_{j,i}$ denote the activation value of channel $i$ at token/sample index $j$. A conservative calibration statistic is the maximum absolute value:
\begin{equation}
    a_i^{\max} = \max_j |x_{j,i}|.
\end{equation}
This statistic avoids clipping calibration observations but can be dominated by a small number of rare outliers. Under W4A4 activation quantization, such tails are costly because the representable values are sparse.

We instead support a high-percentile statistic:
\begin{equation}
    a_i^{p} = Q_p(\{|x_{j,i}|\}_j),
    \label{eq:percentile}
\end{equation}
where $Q_p$ is the empirical $p$-th percentile over the calibration activations of that channel. When statistics are accumulated over multiple calibration records, the same percentile rule is applied to the aggregated activation samples before channel masks are constructed. This gives a single robust range estimate per input channel while avoiding dependence on the partitioning of the calibration set.

Percentile calibration changes the calibration behavior rather than the runtime numerical format. It allows a controlled fraction of rare activation extremes to be clipped, while allocating more effective resolution to the main body of the activation distribution. This is suitable for low-bit video generation because quality metrics are often more sensitive to systematic rounding noise across many tokens and frames than to a few isolated calibration extremes.

\subsection{Channel Mask Construction}
The collected activation statistic is used to build a SmoothQuant-style channel mask. For input channel $i$, let $w_i$ be the maximum absolute weight magnitude over the output dimension:
\begin{equation}
    w_i = \max_o |W_{o,i}|.
\end{equation}
Given activation statistic $a_i$, balancing coefficient $\alpha$, and numerical stabilizer $\epsilon$, the mask is
\begin{equation}
    m_i =
    \frac{w_i^{\alpha}}
    {(a_i + \epsilon)^{1-\alpha}}.
    \label{eq:mask}
\end{equation}
In the conservative variant, $a_i=a_i^{\max}$. In our percentile-calibrated variant, $a_i=a_i^p$. The scaled linear computation is equivalent before quantization:
\begin{equation}
    xW^T = (x \operatorname{diag}(m))
    (W \operatorname{diag}(m)^{-1})^T.
\end{equation}
The benefit appears after quantization: activation and weight ranges become better balanced before HiFloat4 quantize-dequantize is applied.

\subsection{Compact PTQ State}
A naive deployment artifact could duplicate a full transformed copy of the Wan2.2 transformer weights. Instead, we store a compact PTQ state containing the per-layer channel masks and quantization descriptors needed to reproduce the W4A4 model. At inference time, this state is applied to the original floating-point checkpoint, and each affected weight is transformed and quantized under the restored mask. This design reduces storage overhead and keeps the quantized model explicitly tied to the declared base checkpoint.

\section{Experiments}
\subsection{Experimental Setup}
We evaluate the proposed PTQ pipeline on the text-to-video task. The base Wan2.2 model is loaded in BF16 and serves as the reference model. Calibration uses the OpenS2V-5M-derived JSON prompt file~\cite{opens2v}, where each sample contains a prompt under the \emph{cap} field. Since the organizer-required generation pipeline is image-to-video while the evaluation task is prompt-driven text-to-video generation, we provide an explicit blank placeholder image as the image input. This behavior is fixed during both calibration and inference.

The generation protocol uses resolution $720 \times 1280$, $61$ frames, $40$ denoising steps, and classifier-free guidance scale $3.5$. Calibration uses $16$ prompts with random seed $42$. The W4A4 model uses the same base checkpoint, prompt set, placeholder-image condition, resolution, frame count, and sampling steps as the BF16 baseline.

\begin{table}[t]
\caption{Default calibration and generation settings.}
\label{tab:setup}
\centering
\begin{tabular}{lc}
\toprule
Item & Setting \\
\midrule
Base model & Wan2.2-I2V-A14B \\
Precision baseline & BF16 \\
Quantized format & HiFloat4 W4A4 \\
Target modules & Two Wan transformer modules \\
Calibration prompts & 16 \\
Resolution & $720 \times 1280$ \\
Frames & 61 \\
Denoising steps & 40 \\
Guidance scale & 3.5 \\
Seed & 42 \\
\bottomrule
\end{tabular}
\end{table}

\subsection{Implementation Details}
Table~\ref{tab:quant_summary} summarizes the layer conversion. Each transformer has $400$ HiFloat4 quantized linear layers and $6$ high-precision linear layers. Across both transformer modules, $800$ linear layers are quantized. The retained high-precision layers are boundary modules such as time/text embedding projections and output projections. No full transformer block is retained in high precision in the default configuration, leaving the explicit high-precision block budget unused.

\begin{table}[t]
\caption{Layer conversion summary for the default W4A4 configuration.}
\label{tab:quant_summary}
\centering
\begin{tabular}{lccc}
\toprule
Module & HiF4 Linear & FP Linear & FP Blocks \\
\midrule
Transformer-1 & 400 & 6 & 0 \\
Transformer-2 & 400 & 6 & 0 \\
\midrule
Total & 800 & 12 & 0 \\
\bottomrule
\end{tabular}
\end{table}

The overall procedure follows the three stages in Fig.~\ref{fig:pipeline}. First, calibration estimates activation statistics on the BF16 model. Second, PTQ converts the selected projections, applies channel balancing, and records the compact quantization state. Third, inference restores the quantized model from the base checkpoint and the PTQ state before generating videos. This staged design separates statistical calibration, model conversion, and evaluation, making the reported comparison reproducible without changing the sampling protocol.

\subsection{Main Results}
Table~\ref{tab:main_results} reports the main comparison under matched generation and evaluation settings. The HiFloat4 W4A4 model reduces the unweighted mean score from $0.6800$ to $0.5880$, corresponding to a drop of $0.0920$. The largest degradation appears in subject consistency, indicating that aggressive 4-bit activation quantization affects identity and object preservation. In contrast, aesthetic quality and overall consistency remain comparable to the BF16 baseline, and motion smoothness has only a small drop.

\begin{table}[t]
\caption{Main results on the official video generation evaluation. Higher is better for all metrics.}
\label{tab:main_results}
\centering
\footnotesize
\setlength{\tabcolsep}{3.5pt}
\begin{tabular}{lccc}
\toprule
Metric & BF16 & HiFloat4 W4A4 & Delta \\
\midrule
Imaging quality & 0.7027 & 0.6507 & -0.0520 \\
Aesthetic quality & 0.5456 & 0.5458 & +0.0002 \\
Overall consistency & 0.2263 & 0.2308 & +0.0045 \\
Subject consistency & 0.9331 & 0.5324 & -0.4007 \\
Motion smoothness & 0.9923 & 0.9803 & -0.0120 \\
Unweighted mean & 0.6800 & 0.5880 & -0.0920 \\
\bottomrule
\end{tabular}
\end{table}

\subsection{Qualitative Results}
\begin{figure}[t]
\centering
\includegraphics[width=\columnwidth]{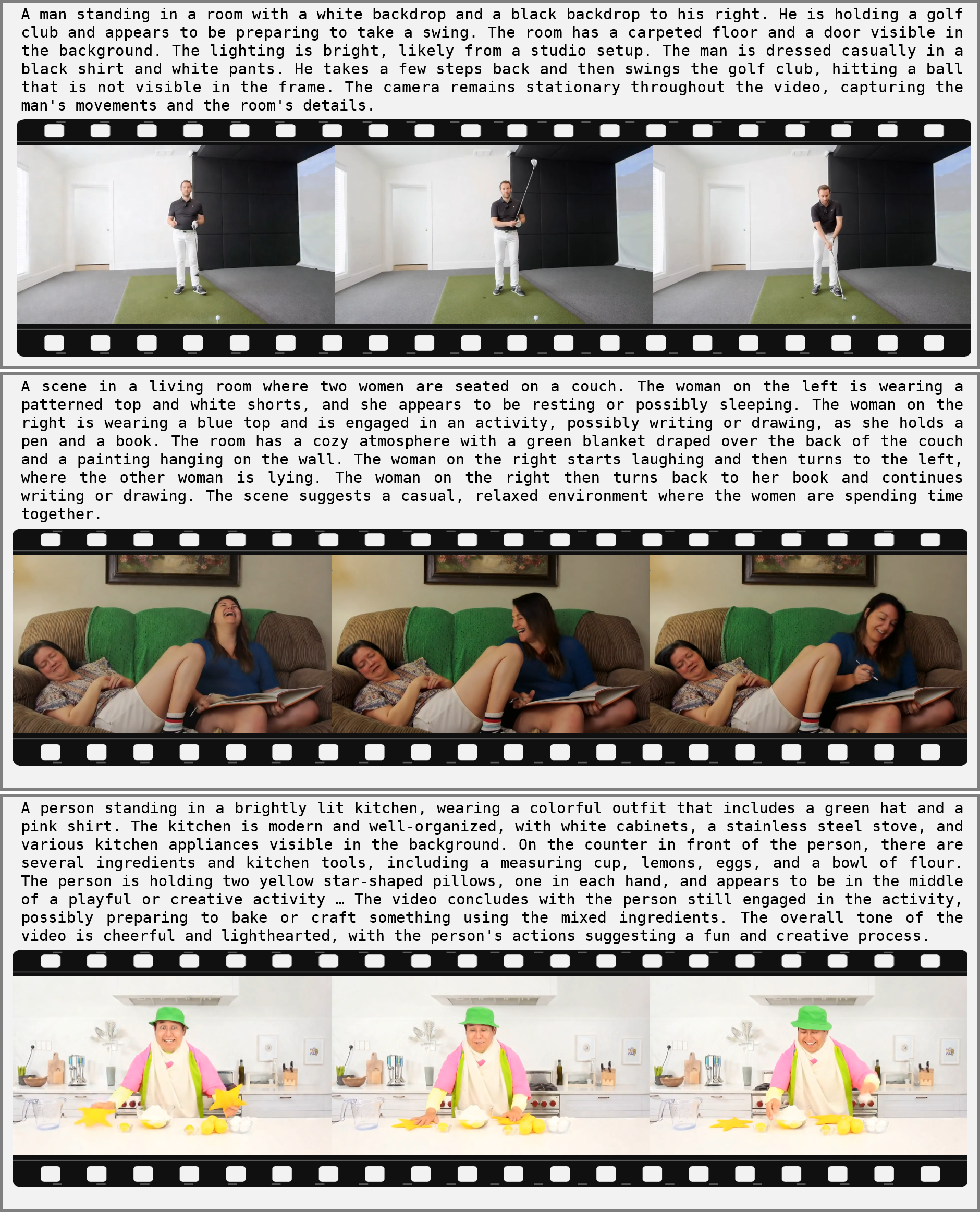}
\caption{Qualitative W4A4 examples. Each row shows the prompt and sampled frames from one generated video.}
\label{fig:qualitative}
\end{figure}

Fig.~\ref{fig:qualitative} shows qualitative W4A4 generations with their prompts and sampled frames. The examples illustrate that the quantized model preserves scene layout and plausible motion, while fine subject details and local object geometry remain sensitive under 4-bit activation quantization, consistent with the subject-consistency drop in Table~\ref{tab:main_results}.

\subsection{Analysis}
The metric profile suggests that the main failure mode of W4A4 Wan2.2 is not temporal smoothness. Motion smoothness remains close to BF16, which indicates that the denoising trajectory still produces temporally coherent frame transitions. Aesthetic quality is also stable, suggesting that global visual style and frame-level appeal are not the most sensitive components under this PTQ configuration.

Subject consistency is the primary bottleneck. This is expected for W4A4 activation quantization because subject identity depends on repeated preservation of semantic and local visual details across frames. Small quantization errors in attention and feed-forward projections can accumulate over denoising steps and appear as subject drift, local blur, texture collapse, or object inconsistency. Percentile calibration is designed to reduce systematic range over-expansion caused by rare activation tails, but it does not eliminate all errors introduced by quantizing both weights and activations to 4 bits.

The compact PTQ state is also important for reproducibility. Since it stores transformation parameters rather than full transformed weights, inference remains tied to the declared base Wan2.2 checkpoint. This makes the base model, quantization configuration, prompt source, resolution, and frame count explicit, and it also reduces the storage overhead of comparing multiple PTQ variants.

\subsection{Limitations}
The method relies on calibration prompts and percentile hyperparameters. If the calibration prompts are too few or distributionally different from the evaluation prompts, the selected percentile may under-cover important activation modes. The current system does not use the optional ViDiT-Q rotation path, so part of the channel imbalance may remain unresolved. Finally, W4A4 quantization remains a severe setting for video generation; preserving subject identity under 4-bit activations is still challenging even when global quality and motion metrics remain stable.

\newpage
\section{Conclusion}
We presented Tail-Aware HiFloat4, a percentile-calibrated W4A4 PTQ pipeline for Wan2.2 text-to-video generation. The method adapts ViDiT-Q-style calibration and channel balancing to the dual-transformer Wan2.2 pipeline, uses HiFloat4 quantize-dequantize for both weights and activations in target linear layers, and stores only compact PTQ deltas for inference. The results show that several global video-quality metrics remain close to the BF16 baseline, while subject consistency is the dominant remaining degradation. Future improvements should focus on identity-sensitive layers, calibration prompt coverage, and selective use of rotation or high-precision retention under the challenge constraints.

\IEEEtriggeratref{13}
\bibliographystyle{IEEEtran}
\bibliography{main}

\end{document}